\documentclass{ieeeaccess}
\usepackage{cite}
\usepackage{amsmath,amssymb,amsfonts}
\usepackage{algorithmic}
\usepackage{graphicx}
\usepackage{textcomp}

\usepackage[utf8]{inputenc} 
\usepackage[T1]{fontenc}    
\usepackage{url}            
\usepackage{booktabs}       
\usepackage{amsfonts}       
\usepackage{nicefrac}       
\usepackage{microtype}      
\usepackage{enumitem}
\usepackage{amsmath}
\usepackage{multirow} 
\usepackage{subcaption}
\usepackage{kotex}
\usepackage{soul}

\usepackage{bm}
\makeatletter
\AtBeginDocument{\DeclareMathVersion{bold}
\SetSymbolFont{operators}{bold}{T1}{times}{b}{n}
\SetSymbolFont{NewLetters}{bold}{T1}{times}{b}{it}
\SetMathAlphabet{\mathrm}{bold}{T1}{times}{b}{n}
\SetMathAlphabet{\mathit}{bold}{T1}{times}{b}{it}
\SetMathAlphabet{\mathbf}{bold}{T1}{times}{b}{n}
\SetMathAlphabet{\mathtt}{bold}{OT1}{pcr}{b}{n}
\SetSymbolFont{symbols}{bold}{OMS}{cmsy}{b}{n}
\renewcommand\boldmath{\@nomath\boldmath\mathversion{bold}}}
\makeatother

\def\BibTeX{{\rm B\kern-.05em{\sc i\kern-.025em b}\kern-.08em
    T\kern-.1667em\lower.7ex\hbox{E}\kern-.125emX}}

\begin{document}
\history{Date of publication 2 July 2025, date of current version 10 July 202}
\doi{10.1109/ACCESS.2025.3585106}

\title{The Role of Teacher Calibration in Knowledge Distillation}
\author{\uppercase{Suyoung Kim}\authorrefmark{1}, 
\uppercase{Seonguk Park}\authorrefmark{2}, \uppercase{Junhoo Lee}\authorrefmark{1}, and \uppercase{Nojun Kwak}\authorrefmark{1}}

\address[1]{Department of Intelligence and Information, Seoul National University,  Gwanak, Seoul 08826, Republic of Korea}
\address[2]{A2Mind, Gangnam, Seoul 06349, Republic of Korea}


\markboth
{S. Kim et al.: The Role of Teacher Calibration in Knowledge Distillation}
{S. Kim et al.: The Role of Teacher Calibration in Knowledge Distillation}


\begin{abstract}
Knowledge Distillation (KD) has emerged as an effective model compression technique in deep learning, enabling the transfer of knowledge from a large teacher model to a compact student model. While KD has demonstrated significant success, it is not yet fully understood which factors contribute to improving the student's performance.
In this paper, we reveal a strong correlation between the teacher's calibration error and the student's accuracy. Therefore, we claim that the calibration of the teacher model is an important factor for effective KD.
Furthermore, we demonstrate that the performance of KD can be improved by simply employing a calibration method that reduces the teacher's calibration error.
Our algorithm is versatile, demonstrating effectiveness across various tasks from classification to detection. Moreover, it can be easily integrated with existing state-of-the-art methods, consistently achieving superior performance.
\end{abstract}


\titlepgskip=-21pt

\maketitle

\section{Introduction}
\label{sec:intro}

\begin{figure*}[t]
    \centering
    \parbox{0.75\textwidth}{
        \centering
        {\includegraphics[width=0.49\linewidth]{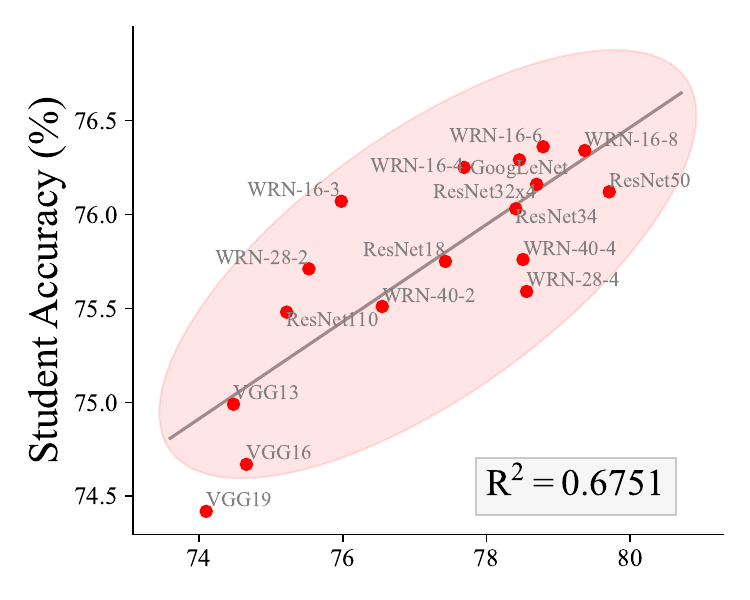}}
        {\includegraphics[width=0.49\linewidth]{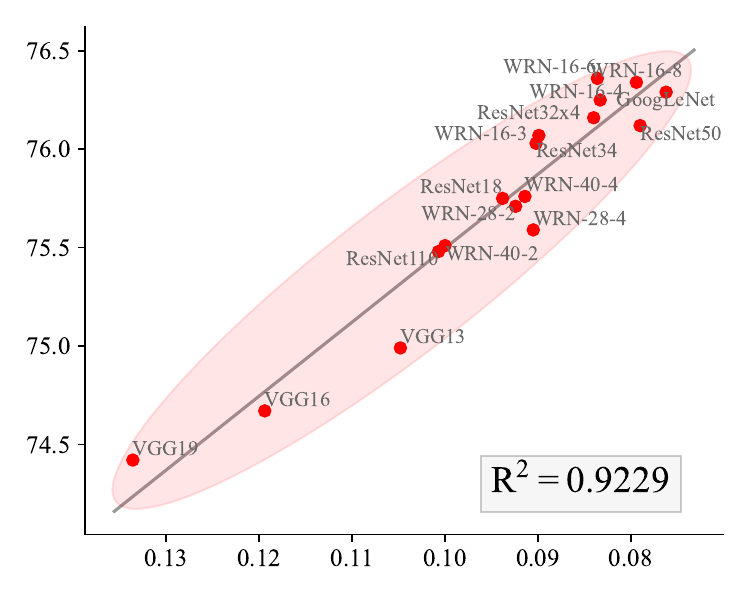}}
    }
    \parbox{0.75\textwidth}{
        \centering
        {\includegraphics[width=0.49\linewidth]{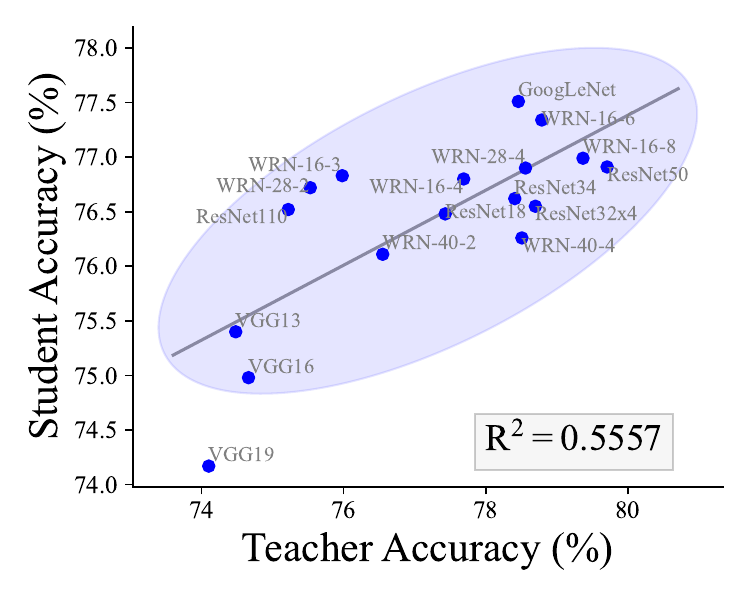}}
        {\includegraphics[width=0.49\linewidth]{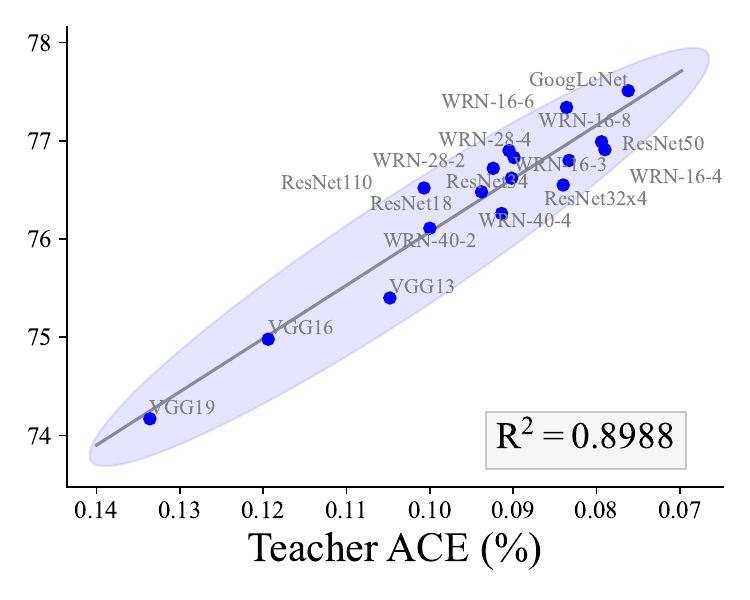}}
    }
    \caption{\textbf{Left: Correlation for Teacher Accuracy and Student Accuracy. Right: Teacher Adaptive Calibration Error (ACE) and Student Accuracy.} The figure presents the outcomes of Knowledge Distillation (KD) training using 17 distinct pretrained teacher models. Each point indicates the teacher used for KD. Two student architectures are examined: WRN-16-2 (\textcolor{red}{top}) and ShuffleNetV2 (\textcolor{blue}{bottom}). All experiments are conducted on the CIFAR-100 dataset. Note that the `teacher's ACE' is more correlated with the `student's accuracy' than the `teacher's accuracy'.}
    \label{fig:ace_sacc}
\end{figure*}

With the recent emergence of diverse applications of deep neural networks, extensive research has been conducted on techniques to compress deep neural networks.
Among them, knowledge distillation (KD) is a model training strategy that boosts the performance of a smaller model, so that it can replace a larger model. 
The goal of KD is to successfully transfer knowledge from a larger network, known as the teacher, to a smaller network, called the student.
This process naturally yields two design criteria of KD:
First, the choice of the distillation method used to transfer the knowledge of the teacher to the student, and second, the choice of an appropriate teacher for the student network.

The majority of contemporary works explore state-of-the-art (SOTA) distillation methods using fixed sets of teacher-student pairs. However, studying the criterion of choosing an appropriate teacher is important as well, as the simultaneous consideration of the two design criteria can ultimately provide a more optimized solution. For example, recent studies \cite{cho2019efficacy,zong2022better,beyer2022knowledge} show that a high-accuracy teacher does not guarantee better performance of a student in knowledge distillation, and these findings highlight the need for methods of assessing a teacher and identifying a `good teacher.' 
Our paper focuses on this issue. By analyzing the learning dynamics of deep learning empirically, we provide a high-level explanation about the question, "Which factor of teacher induces effective KD." 
As an answer, we demonstrate that the calibration error of the teacher plays an important role, and significant performance improvements can be achieved in the KD process by simply applying the calibration method to the teacher network.
We found that not only the standard KD but also the highly fine-tuned state-of-the-art KD method that has been designed in a sophisticated manner can benefit from calibration method.

To delve into the details of our proposed method,
we first propose using the calibration error as a criterion to evaluate the performance of the teacher.
In statistics, calibration refers to the process of adjusting a probability model to ensure that its predictions closely align with actual probability. For instance, if the model predicts output with a probability (confidence) of 70\% for a set of particular data samples, then ideally, 70\% of those predictions should be correct to be considered well-calibrated. In contrast, a poorly calibrated model might predict an outcome with, for example, a 99\% probability, even though only 70\% of those predictions are correct. In this case, the model is said to be overconfident. Poorly calibrated models, like this case, can be problematic in real-world applications. For example, misplaced confidence can lead to severe consequences in medical diagnostics or autonomous vehicle decisions. Therefore, reducing calibration error is crucial for these applications.

When it comes to deep learning models, \cite{guo2017calibration} raised the issue that calibration error tends to increase as models become increasingly complex and achieve higher performance. As a result, follow-up studies have actively analyzed characteristics of calibration error in deep learning models \cite{nixon2019measuring}. In our research, we demonstrate that a well-calibrated model is not only probabilistically reliable but also effective when used in KD.

In Figure~\ref{fig:ace_sacc}, our experiment reveals a significant correlation between the teacher model's calibration error and the student model's accuracy in the knowledge distillation training. These observations lead us to conclude that the teacher model's calibration error is a critical factor in the effectiveness of knowledge distillation.

Furthermore, our paper demonstrates that reducing the teacher model's calibration error can improve the performance of KD. By applying a simple temperature-based calibration method to teachers, we achieve consistent performance improvement against state-of-the-art models. Our experiments provide empirical evidence that the calibration error plays a crucial role in KD.

In summary, the main contributions are as follows:
\begin{itemize}[leftmargin=*] 
    \item We empirically show that the calibration error of the teacher negatively correlates with the performance of the student network. These experimental results demonstrate that the calibration error of the teacher is a significant factor in KD.

    \item We demonstrate that the performance of knowledge distillation can be significantly improved by merely applying a simple calibration method to the teacher model.
    By applying this simple calibration to classification and detection tasks, we consistently achieve superior results across various experimental settings.
\end{itemize}
\section{Related Work}
\label{sec:related}

\subsection{Knowledge Distillation} 
Knowledge distillation is one of the deep learning compression techniques introduced by \cite{hinton2015kd}. It leverages information from a larger, high-performing teacher model to train a smaller student model. By training students using KD, students can achieve better performance without additional memory and computational complexity.

Most KD papers proposed techniques to boost student performance.
FitNet \cite{romero2014fitnets} and FT \cite{NEURIPS2018_6d9cb7de} utilized auxiliary networks called `regressor' to assist the process of feature-map distillation.
Methods such as DML \cite{zhang2018deep} and AFD \cite{pmlr-v119-chung20a} proposed online distillation methods that train the teacher and student simultaneously.
TAKD \cite{mirzadeh2020takd} and DGKD \cite{Son_2021_ICCV} proposed utilizing `teacher assistants' that help bridge the gap between the student and the teacher.
Recently, methods such as MLLD \cite{jin2023mlld} and ReviewKD \cite{chen2021reviewkd} set SOTA performance in logit distillation and feature distillation, respectively.

However, with numerous distillation methods that try to achieve so-called SOTA performance, the reason why each method should perform better than previous literature still remains very obscure. Thus, many studies discussed the fundamental reasons for the performance enhancements of KD.
\cite{furlanello2018born_again} (BAN), and \cite{zhao2022dkd} (DKD) analyzed the effect of label smoothing and negative logits on KD. 
\cite{cho2019efficacy} showed that the better teacher did not promise the better performance of the student network and provided a solution based on early-stopping. IPWD \cite{niu2022respecting} showed the importance of bridging the classwise knowledge imbalance gap between the teacher and the student. \cite{zhou2021rethinking, menon2021statistical} speculated the role of KD based on the bias-variance trade-offs. 
\cite{yang2021rethinking} deals with overconfident problems in KD, which is the closest approach to ours. However, their analysis and experiments are constrained to a limited scope. 
Our work following the findings of \cite{cho2019efficacy}, we inspect the comprehensive role of the teacher's `calibration error' on knowledge distillation.

\subsection{Deep learning calibration}

Model calibration refers to making the model’s actual accuracy reflect the actual confidence or vice-versa.
Since the machine learning era, lots of studies have been conducted to decrease the calibration error of a model. Some representative examples are Histogram binning \cite{zadrozny2001histogram}, Isotonic regression \cite{zadrozny2002isotonic}, Platt scaling \cite{platt1999platt}, vector scaling, and temperature scaling. Rather recently, \cite{guo2017calibration} reported that simply using temperature scaling calibrates well enough for deep learning models. Following the work of \cite{guo2017calibration}, we investigate the usage of temperature scaling on the DNNs, which doesn't change the model accuracy and doesn't have additional costs for training or inference.

\section{Method}
\label{sec:method}

\subsection{Metric for calibration error}

Expected Calibration Error (ECE) is the metric for measuring the calibration error of a certain probabilistic model. The ECE value can be acquired as follows:
\begin{equation}
    \text{ECE} = \sum_{m=1}^{M} \frac{|B_m|}{N} \left| \text{acc}(B_m) - \text{conf}(B_m) \right|,
    \label{eq:ece}
\end{equation}
where $N$ is the number of samples, $M$ is the number of bins dividing the accuracy space of [0,1] evenly,  $B_m$ is the set of samples belonging to interval $m$, acc($B_m$) refers to the accuracy of interval $m$, and conf($B_m$) refers to the average prediction probability (confidence) of samples in interval $m$.

To identify the source of a model's calibration error, we can decompose the ECE into its components: overconfident ECE ($\text{ECE}_{o}$) and underconfident ECE ($\text{ECE}_{u}$). The decomposed ECE can be formulated as follows:
\begin{equation}
    \text{ECE}_{o} = \sum_{m=1}^{M} \frac{|B_m|}{N}  \max(\text{conf}(B_m) - \text{acc}(B_m), 0),
    \label{eq:ece_over}
\end{equation}
\begin{equation}
    \text{ECE}_{u} = \sum_{m=1}^{M} \frac{|B_m|}{N}  \max(\text{acc}(B_m) - \text{conf}(B_m), 0).
    \label{eq:ece_under}
\end{equation}
%
%
$\text{ECE}_{o}$ represents the sum of the overconfident components of the calibration error, while $\text{ECE}_{u}$ captures the sum of the underconfident components. These two metrics allow us to determine whether the calibration error of a model predominantly originates from its overconfident or underconfident predictions.

Originally, ECE was designed for binary classification~\cite{degroot1983ece} 
and has been adapted for multi-class problems by treating the problem with $K$ classes as a set of binary predictions: one for the target label and the other for non-target labels. However, this approach neglects the information that deep learning models may capture between non-target classes. Such ignorance is 
problematic for knowledge distillation, which inherently involves transferring information about inter-class relationships \cite{furlanello2018born_again}. Additionally, ECE suffers from another limitation in that it creates bins by evenly spacing the probability space. This trait results in scenarios where fewer samples at
lower probability levels contribute significantly to the ECE. Thus, ECE may not be the most effective metric for measuring the calibration of deep learning models that often display highly overconfident probability distributions.

Given these limitations, this paper employs Adaptive Calibration Error (ACE)~\cite{nixon2019measuring} as a metric for measuring calibration error. ACE accounts for multi-class predictions and employs adaptive binning to equalize the number of samples in each bin. The formula for ACE is as follows:
\begin{equation}
    \text{ACE} = \frac{1}{KR}\sum_{k=1}^{K}\sum_{r=1}^{R}  \left| \text{acc}(r,k) - \text{conf}(r,k) \right|.
    \label{eq:ace}
\end{equation}
%
%
Here, $K$ represents the number of classes, and $R$ signifies the number of bins. The terms acc$(r, k)$ and conf($r, k$) refer to the accuracy and confidence (output probability), respectively, of the samples belonging to class $k$ in the $r$-th bin. The calibration range $r$ is defined by the  $\lfloor N/R \rfloor$th index, where $N$ is the number of samples. ACE offers several advantages over ECE. Specifically, it computes errors for non-target probabilities and allocates an equal number of samples to each bin. These features make it especially well-suited for assessing the calibration of deep learning models when the probability distribution is skewed.

\subsection{Calibration error and KD performance}
\label{subsec:criterion_ace}

In the previous section, we highlighted that the accuracy of the teacher is not the best indicator for the better accuracy of the student model, proposing the calibration error as a more reliable metric. To validate this hypothesis, we examined the correlation between the accuracy of the student model and both the accuracy and calibration error of the teacher model.
Figure~\ref{fig:ace_sacc} is the standard KD training result using 17 different teacher models for fixed students.
Figure~\ref{fig:ace_sacc} clearly illustrates a strong correlation between the teacher's calibration error and the student models' accuracy. 
The $R^2$ values for the correlation between the teacher model's calibration error and the student model's accuracy are 0.9229 for WRN-16-2 and 0.8998 for ShuffleNetV2, respectively. 
The accuracies of the teacher models show relatively lower correlations with the student's accuracy, where the values are only 0.6751 and 0.5557.
These experimental results underscore the need to consider calibration error as a key factor for effective KD.

These tendencies suggest that teachers with lower calibration errors perform better in knowledge distillation. 
Teachers with lower calibration errors offer two distinct advantages compared to those with higher errors.
First, they provide a more reliable basis for Kullback-Leibler (KL) divergence to take effect in KD. KL divergence fundamentally measures the difference between two probability distributions, and in this context, better calibration allows the teacher's output to form a more accurate and reliable probability distribution. Essentially, calibration error measures how well the model's output probability aligns with true accuracy; hence, a lower calibration error means the model forms a better mathematical probability distribution with respect to the input distribution.
Second, teachers with lower calibration errors act as stronger regularizers. The ground truth label is essentially a label with 100\% confidence, and teachers with lower calibration errors (i.e., less overconfident teachers) play a larger role as label smoothers for the true label. This amplifies one of the benefits of KD, which takes its role as a regularize.
If we set teacher output probability of class $i$ as $p^{(i)}$ and student output probability $q^{(i)}$, we can decompose the teacher probability to calibrated probability, and error probability. Then $p^{(i)} = (1-k)p^{(i)}_{\mathrm{cal}} + kp^{(i)}_{\mathrm{error}}$ when $k\in[0,1]$ represent the intensity of overconfident error. Since overcalibrated error can be expressed by one hot vector, we can approximate $p^{(i)}_{\mathrm{error}} \approx y^{(i)}$.
Then the KD loss becomes

\begin{align}
L_\mathrm{Total}^{(i)}
  &= (1-\lambda)\,H\!\left(y^{(i)},q^{(i)}\right) +\lambda\,D_{\mathrm{KL}}\!\left(p^{(i)},q^{(i)}\right) \\[4pt]
  &= -(1-\lambda)\,y^{(i)}\log q^{(i)}- \lambda\,p^{(i)}\log q^{(i)} + C \\[4pt]
  &= -(1-\lambda)\,y^{(i)}\log q^{(i)} \nonumber \\
  &\quad - \lambda\bigl[(1-k)\,p^{(i)}_{\text{cal}}
           + k\,p^{(i)}_{\text{error}}\bigr]\log q^{(i)} \\[4pt]
&= -\underbrace{\bigl(1-\lambda+\lambda k\bigr)\,y^{(i)}\log q^{(i)}}_{\text{One-hot}}
   \;-\;
   \underbrace{\lambda(1-k)\,p^{(i)}_{\text{cal}}\log q^{(i)}}_{\text{KD}}.
\end{align}

We ignore constant factor C since it doesn't affect to optimization process. After decomposing calibrated and overconfident probability, we find that the coefficient associated with the overconfident error diminishes the influence of the KD loss. 
Therefore, it is imperative to find non-overconfident teachers for successful KD.

\subsection{Enhancing KD with calibration method}

KD is a process where the teacher conveys the probability simplex to the student, allowing the student to learn the relationships and uncertainties between classes.
However, there arises a question of whether this process can truly be performed successfully. Typically, cross-entropy (CE) or its variations are used for training the model, and one of the characteristics of these losses is that they can't reach zero, pushing the value of the highest-valued logit larger and larger, thus producing overconfident outputs \cite{svm_jh}.

Given these challenges, utilizing teachers that are not overly confident becomes crucial for successful KD. 
We have discussed the overconfidence prior present in deep learning models. Regarding calibration error functions as a metric to assess the alignment between a model's probability (confidence) distribution and the accuracy distribution, overconfident prior produces a high calibration error. If we can address this issue, we can anticipate an improvement in the model's calibration error.

The overconfidence issue arises when the logit value of the predicted class is excessively high while the remaining logit values are substantially lower. We resolve this issue with a straightforward yet effective solution: incorporating temperature scaling into the teacher's output logit softmax calculation to smooth the probability distribution. As a result of this adjustment, the confidence distribution of the teacher model becomes smoother, thereby alleviating the overconfidence issue. In this paper, we demonstrate that this strategy can significantly enhance the baseline performance. Furthermore, this method proves to be effective not only with standard KD approaches but also with finely tuned state-of-the-art KD techniques.
\section{Experiments}
\label{sec:experiments}

\subsection{Implementation detail}

\begin{table*}[ht]
    \centering
        \caption{\textbf{Results on the CIFAR-100 dataset with homogenous Teacher-Student architectures.} The table presents the experimental results of various state-of-the-art KD methods, including feature-based and logit-based distillation methods. $\dagger$ represents the result implemented by ours which is the average of three trials, and the standard deviation is expressed within parentheses.}

    \resizebox{0.85\textwidth}{!}{%
    \begin{tabular}{@{}cl ccccccc@{}}
        \toprule
        \multirow{4}{*}{Method} &
        \multicolumn{1}{c}{\multirow{2}{*}{Teacher}} &
        ResNet56 & ResNet110 & ResNet32x4 & WRN-40-2 & WRN-40-2 & VGG13 \\
        & & 72.34 & 74.31 & 79.42 & 75.61 & 75.61 & 74.64 \\
        \cmidrule(lr){2-8}
        & \multicolumn{1}{c}{\multirow{2}{*}{Student}} &
        ResNet20 & ResNet32 & ResNet8x4 & WRN-16-2 & WRN-40-1 & VGG8 \\
        & & 69.06 & 71.14 & 72.50 & 73.26 & 71.98 & 70.36 \\
        \midrule
        \multirow{4}{*}{Features} & FitNet    & 69.21 & 71.06 & 73.50 & 73.58 & 72.24 & 71.02 \\
                                  & CRD       & 71.16 & 73.48 & 75.51 & 75.48 & 74.14 & 73.94 \\
                                  & ReviewKD  & 71.89 & 73.89 & 75.63 & 76.12 & 75.09 & 74.84 \\
        \midrule
        \multirow{5}{*}{Logits}   & $\text{KD}^\dagger$& 70.90 (0.18) & 73.62 (0.34) & 75.69 (0.16) & 75.33 (0.18) & 73.43 (0.37) & 73.96 (0.20) \\
                                  & $\text{KD + Ours}^\dagger$      & \textbf{71.38} (0.32) & \textbf{74.00} (0.17) & \textbf{76.00} (0.12) & \textbf{75.80} (0.13) & \textbf{74.42} (0.12) & \textbf{74.09} (0.15) \\
                                  & $\Delta$ & \textcolor{red}{+0.48} & \textcolor{red}{+0.38} & \textcolor{red}{+0.31} & \textcolor{red}{+0.47} & \textcolor{red}{+0.99} & \textcolor{red}{+0.13} \\
                                  \cmidrule(lr){2-8}
                                  & $\text{MLLD}^\dagger$ & 72.05 (0.25) & 74.48 (0.35) & 77.02 (0.16) & 76.47 (0.16) & 75.56 (0.19) & 74.99 (0.25) \\
                                  & $\text{MLLD + Ours}^\dagger$ & \textbf{72.46} (0.27) & \textbf{74.71} (0.14) & \textbf{77.23} (0.15) & \textbf{76.88} (0.12) & \textbf{76.01} (0.28) & \textbf{75.21} (0.13) \\
                                  & $\Delta$ & \textcolor{red}{+0.41} & \textcolor{red}{+0.23} & \textcolor{red}{+0.21} & \textcolor{red}{+0.41} & \textcolor{red}{+0.45} & \textcolor{red}{+0.22} \\

        \bottomrule
    \end{tabular}%
    }
\label{tab:cifar100_same_architecture}
\end{table*}

\begin{table*}[ht]
    \centering
        \caption{\textbf{Results on the CIFAR-100 dataset with heterogenous Teacher-Student architectures.} The table presents the experimental results of various state-of-the-art KD methods, including feature-based and logit-based distillation methods. $\dagger$ represents the result implemented by ours which is the average of three trials, and the standard deviation is expressed within parentheses.}

    \resizebox{0.85\textwidth}{!}{%
    \begin{tabular}{@{}cl cccccc@{}}
        \toprule
        \multirow{4}{*}{Method} &
        \multicolumn{1}{c}{\multirow{2}{*}{Teacher}} &
        ResNet32x4 & WRN-40-2 & VGG13 & ResNet50 & ResNet32x4 \\
        & & 79.42 & 75.61 & 74.64 & 79.34 & 79.42 \\
        \cmidrule(lr){2-7}
        & \multicolumn{1}{c}{\multirow{2}{*}{Student}} &
        ShuffleNet-V1 & ShuffleNet-V1 & MobileNet-V2 & MobileNet-V2 & ShuffleNet-V2 \\
        & & 70.50 & 70.50 & 64.60 & 64.60 & 71.82 \\
        \midrule
        \multirow{4}{*}{Features} & FitNet & 73.59 & 73.73 & 64.14 & 63.16 & 73.54 \\
                                  & CRD & 75.11 & 76.05 & 69.73 & 69.11 & 75.65 \\
                                  & ReviewKD & 77.45 & 77.14 & 70.37 & 69.89 & 77.78 \\
        \midrule
        \multirow{5}{*}{Logits} & $\text{KD}^\dagger$ & 72.69 (0.13) & 73.13 (0.13) & 65.07 (0.45) & 64.78 (0.34) & \textbf{76.15} (0.13) \\
                                 & $\text{KD + Ours}^\dagger$ & \textbf{73.73} (0.15) & \textbf{73.53} (0.31) & \textbf{67.24} (0.37) & \textbf{66.71} (0.41) & 75.99 (0.23) \\
                                 & $\Delta$ & \textcolor{red}{+1.04} & \textcolor{red}{+0.40} & \textcolor{red}{+2.17} & \textcolor{red}{+1.93} & \textcolor{blue}{-0.16} \\
                                 \cmidrule(lr){2-7}
                                 & $\text{MLLD}^\dagger$ & 77.13 (0.15) & 77.26 (0.20) & 69.70 (0.79) & 69.60 (0.64) & 78.42 (0.19) \\
                                 & $\text{MLLD + Ours}^\dagger$ & \textbf{77.40} (0.21) & \textbf{77.70} (0.31) & \textbf{70.90} (0.22) & \textbf{71.04} (0.18) & \textbf{78.65} (0.15) \\
                                 & $\Delta$ & \textcolor{red}{+0.27} & \textcolor{red}{+0.44} & \textcolor{red}{+1.20} & \textcolor{red}{+1.44} & \textcolor{red}{+0.23} \\

        \bottomrule
    \end{tabular}%
    }

    \label{tab:cifar100_different_architecture}
\end{table*}

There are many calibration methods such as Platt scaling~\cite{platt1999platt}, isotonic regression~\cite{zadrozny2002isotonic}, matrix scaling, vector scaling, and temperature scaling~\cite{guo2017calibration}. We utilized temperature scaling, a simple yet effective calibration method. Temperature scaling is a suitable method for experimentation for some reasons: it does not alter accuracy while modifying the calibration error and does not require an additional validation set. Therefore, we can exclusively reduce the overconfident calibration error while fixing the accuracy. This enables us to investigate the impact of calibration errors on KD. Based on our empirical results, a similar performance boost occurs in KD even when using different calibration methods.
Eq.~\ref{eq:temperature_scaling} below shows the temperature scaling formulation:
\begin{equation}
\hat{p}^{(i)} = \frac{{e^{z^{(i)}/T}}}{{\sum_{j}^{K} e^{z^{(j)}/T}}},
\label{eq:temperature_scaling}
\end{equation}

where $K$ represents the number of classes and $T$ is the temperature parameter,  and $z \in \mathbb{R}^{K}$ denotes the output logit vector of the model, whose $i$-th element being denoted as $z^{(i)}$. Setting $T=1$ makes the equation equivalent to the standard softmax function, and when $T>1$, it softens the output probability. Importantly, changing the temperature does not affect the order of the output probability, meaning that it does not influence the model's prediction accuracy. Thus, changing the temperature gives a way of changing the model's confidence with fixed accuracy. We empirically observed robust performance improvements when the temperature value T was set between 1.5 and 3, indicating that the calibration method enhances KD performance without being sensitive to the exact value of T. Unless otherwise specified in the subsequent experiments, we set the default hyperparameter to $T=1.5$.

We should note that temperature scaling for calibration is different from the one used in standard KD. In standard KD, temperature scaling is typically applied to both the teacher and the student logits. The purpose of temperature scaling in standard KD is to make the student's logit follow the teacher's logit effectively, even for non-target labels.
In contrast, the purpose of temperature scaling in calibration is to enable the student to learn from a well-calibrated teacher. In our experiments, we utilize both temperature scaling techniques. We apply temperature scaling solely to the teacher to reduce the overconfident calibration error and also apply temperature scaling to the teacher and student for better non-target logit distillation. 

After applying temperature scaling to the pre-trained teacher model, the remaining training steps follow the general knowledge distillation process as described by \cite{hinton2015kd}, which uses the cross-entropy loss with the true label and the KL divergence loss with the teacher output.

\subsubsection{CIFAR-100}

In our experiments, the training scheme for KD was configured based on a paper proposed by \mbox{\cite{zhao2022dkd}}. We set the batch size to 128 and conducted training over a total of 240 epochs. The initial learning rate was configured at 0.5 and was decayed by a factor of 10 at epochs 150, 180, and 210. The SGD optimizer was used with a weight decay set to 5e-5 and a momentum of 0.9. For data augmentation, we applied RandomResizedCrop(size=32) and RandomHorizontalFlip(p=0.5) following the previous baseline. The temperature parameter for KD was set to 4, and the loss function combined the cross-entropy with the true labels and the Kullback-Leibler divergence with the teacher's output with student output. The weights for these loss components were set at 0.1 and 0.9, respectively. The primary difference between the standard KD and our enhanced KD scheme lies in the application of temperature scaling to reduce overconfident calibration in the teacher model. For calibration purposes, a temperature of 1.5 was uniformly used across all KD + Ours experiments. For the implementation of MLLD + Ours, we strictly followed the training scheme of MLLD as described in \mbox{\cite{jin2023mlld}}. However, we applied temperature scaling for calibration only when using the KL-divergence loss, one of the four types of losses (Cross entropy loss, KL-divergence loss, batch-level loss, class-level loss) employed in MLLD.

\subsubsection{ImageNet}

For KD training on ImageNet, we followed the training scheme proposed in \mbox{\cite{zhao2022dkd}}. The batch size was set to 512, and the training was conducted over 100 epochs. The initial learning rate was set to 0.2 and was divided by 10 at epochs 30, 60, and 90. We used the SGD optimizer with a weight decay of 1e-4 and a momentum of 0.9. For data augmentation, we applied RandomResizedCrop(size=224) and RandomHorizontalFlip(p=0.5). Similar to CIFAR-100, we employed cross-entropy loss with the true labels and KL-divergence loss with the teacher output, assigning equal weights of 0.5 to each loss. The temperature for KD was set to 1. The only difference between KD and KD + Ours was the application of temperature scaling to calibrate the teacher model.
For the MLLD + Ours experiment, we adhered to the training scheme of MLLD as described in \mbox{\cite{jin2023mlld}}. We applied temperature scaling to the teacher's output for calibration, performed probability smoothing, and then computed the loss.

\subsubsection{MS-COCO object detection}

The object detection experiment was conducted following the training settings for object detection outlined in a previous knowledge distillation baseline paper~\mbox{\cite{chen2021reviewkd}}. We carried out our experiments using the COCO2017 dataset and evaluated it with a validation set. The training was conducted for a total of 90,000 iterations, with the learning rate initially set to 0.02 and reduced by a factor of 10 at the 60,000th and 80,000th iterations. We used the SGD optimizer with a momentum of 0.9 and set the batch size to 16. For the teacher models in our experiments, we used pretrained models provided by the open-source detection library Detectron2~\mbox{\cite{wu2019detectron2}}. The application of our method involved applying temperature scaling to the R-CNN classifier loss of the teacher model in the two-stage object detector, Faster R-CNN. All of the above experiments were conducted using an NVIDIA A100 80GB GPU.

\subsection{Making a better teacher via calibration method} 
\label{subsec:calibrated_kd}
\begin{table}[t]
\caption{Top-1 and top-5 accuracy (\%) on the ImageNet validation. We set ResNet-34 as the teacher and ResNet-18 as the student. $\dagger$ represents the result implemented by ours.}
\resizebox{\columnwidth}{!}{%
\begin{tabular}{@{}llllll@{}}
    \toprule
                          &             & Top-1 & Top-5 & Top-1 & Top-5 \\
    \cmidrule(lr){3-4} \cmidrule(lr){5-6}
    \multirow{4}{*}{Methods} & \multirow{2}{*}{Teacher} & \multicolumn{2}{c}{ResNet34} & \multicolumn{2}{c}{ResNet50}     \\
                          &             & 73.31 & 91.42 & 76.16 & 92.86 \\
    \cmidrule{2-6}
                         & \multirow{2}{*}{Student} & \multicolumn{2}{c}{ResNet18} & \multicolumn{2}{c}{MobileNet-V1} \\
                          &             & 69.75 & 89.07 & 68.87 & 88.76 \\
    \midrule
    \multirow{4}{*}{Features} & AT          & 70.69 & 90.01 & 69.56 & 89.33 \\
                          & CRD         & 71.17 & 90.13 & 71.37 & 90.41 \\
                          & ReviewKD    & 71.61 & 90.51 & 72.56 & 91.00 \\
    \midrule
    \multirow{6}{*}{Logits}   & $\text{KD}^\dagger$          & 70.66 & 89.88 & 68.58 & 88.98 \\
                          & $\text{KD + Ours}^\dagger$   & \textbf{71.60} & \textbf{90.24} & \textbf{71.55} & \textbf{90.43} \\
                          & DKD         & 71.70 & 90.41 & 72.05 & 91.05 \\
                          & $\text{MLLD}^\dagger$        & 71.60 & 90.68 & 73.05 & 91.34 \\
                          & $\text{MLLD + Ours}^\dagger$ & \textbf{71.90} & \textbf{90.72} & \textbf{73.09} & \textbf{91.43} \\
    \bottomrule
\end{tabular}
}
 \label{tab:imagenet_both}
\end{table}
\begin{table*}[ht]
\centering
\caption{\textbf{Experiment results on MS-COCO object detection task.} 
This table shows the result that our method also works effectively for object detection.}
\label{tab:detection}
\resizebox{0.99\textwidth}{!}{%
\begin{tabular}{@{}lcccccc|cccccc|cccccc@{}}
\toprule
& \multicolumn{6}{c}{ResNet101 - ResNet50} & \multicolumn{6}{c}{ResNet101 - ResNet18} & \multicolumn{6}{c}{ResNet50 - MobileNetV2} \\
\cmidrule(lr){2-7} \cmidrule(lr){8-13} \cmidrule(lr){14-19}
& mAP & AP50 & AP75 & APl & APm & APs & mAP & AP50 & AP75 & APl & APm & APs & mAP & AP50 & AP75 & APl & APm & APs \\
\midrule
Teacher & 42.04 & 62.48 & 45.88 & 54.60 & 45.55 & 25.22 & 42.04 & 62.48 & 45.88 & 54.60 & 45.55 & 25.22 & 37.93 & 58.84 & 41.05 & 49.10 & 41.14 & 22.44 \\
Student & 37.93 & 58.84 & 41.05 & 49.10 & 41.14 & 22.44 & 33.26 & 53.61 & 35.26 & 43.16 & 35.68 & 18.96 & 29.47 & 48.87 & 30.90 & 38.86 & 30.77 & 16.33 \\
\midrule
KD & 38.35 & 59.41 & 41.71 & 49.48 & 41.80 & 22.73 & 33.97 & 54.66 & 36.62 & 44.14 & 36.67 & 18.71 & 30.13 & 50.28 & 31.35 & 39.56 & 31.91 & 16.69 \\
KD+Ours & \textbf{39.04} & \textbf{60.74} & \textbf{42.21} & \textbf{50.38} & \textbf{42.35} & \textbf{22.88} & \textbf{34.65} & \textbf{55.99} & \textbf{36.90} & \textbf{45.30} & \textbf{37.26} & \textbf{20.00} & \textbf{31.90} & \textbf{52.81} & \textbf{33.49} & \textbf{41.23} & \textbf{34.05} & \textbf{18.29} \\
$\Delta$ & \textcolor{red}{+0.69} & \textcolor{red}{+1.33} & \textcolor{red}{+0.50} & \textcolor{red}{+0.90} & \textcolor{red}{+0.55} & \textcolor{red}{+0.15} & \textcolor{red}{+0.68} & \textcolor{red}{+1.33} & \textcolor{red}{+0.28} & \textcolor{red}{+1.16} & \textcolor{red}{+0.59} & \textcolor{red}{+1.29} & \textcolor{red}{+1.77} & \textcolor{red}{+2.53} & \textcolor{red}{+2.14} & \textcolor{red}{+1.67} & \textcolor{red}{+2.14} & \textcolor{red}{+1.60} \\
\midrule
ReviewKD & 40.36 & 60.97 & 44.08 & 52.87 & 43.81 & 23.60 & 36.75 & 56.72 & 39.00 & \textbf{49.58} & 39.51 & 19.42 & 33.71 & 53.15 & 36.13 & \textbf{46.47} & 35.81 & 16.77 \\
ReviewKD+Ours & \textbf{40.83} & \textbf{61.82} & \textbf{44.32} & \textbf{53.09} & \textbf{44.22} & \textbf{24.19} & \textbf{36.76} & \textbf{57.27} & \textbf{39.46} & 49.02 & \textbf{39.59} & \textbf{19.86} & \textbf{34.14} & \textbf{54.53} & \textbf{36.48} & 45.80 & \textbf{36.65} & \textbf{17.96} \\
$\Delta$ & \textcolor{red}{+0.47} & \textcolor{red}{+0.85} & \textcolor{red}{+0.24} & \textcolor{red}{+0.22} & \textcolor{red}{+0.41} & \textcolor{red}{+0.59} & \textcolor{red}{+0.01} & \textcolor{red}{+0.55} & \textcolor{red}{+0.46} & \textcolor{blue}{-0.56} & \textcolor{red}{+0.08} & \textcolor{red}{+0.44} & \textcolor{red}{+0.43} & \textcolor{red}{+1.38} & \textcolor{red}{+0.35} & \textcolor{blue}{-0.67} & \textcolor{red}{+0.84} & \textcolor{red}{+1.19} \\
\bottomrule
\end{tabular}
}
\vspace{-2mm}
\end{table*}

\label{sec:ablation_on_temperature}

\begin{figure}[t]
    \centering
    \includegraphics[width=\columnwidth]{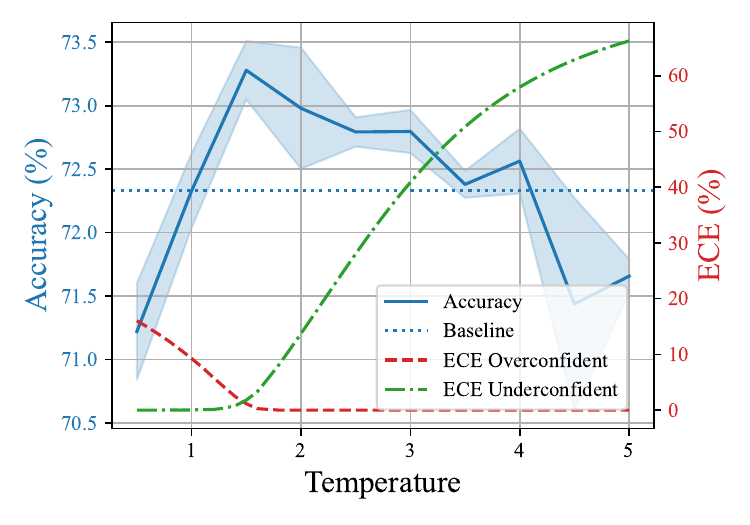}
    \caption{\textbf{Effect of temperature of teacher.} The figure illustrates the impact of varying the temperature parameter on the performance of student models. The teacher-student architecture consists of the ResNet32x4 and WRN-16-2 pair. Experiments were conducted on the CIFAR-100 dataset, with each experiment being run three times to calculate the mean and standard deviation.}
    \label{fig:temperature_ablation}
\end{figure}


In our experiments, we investigated the potential of calibration methods that enhance the performance of knowledge distillation. Our findings demonstrate that a teacher model calibrated through temperature scaling improves KD performance. Our experiment also shows higher KD performance when operated in a slightly underconfident state through higher temperature settings. Figure~\ref{fig:temperature_ablation} provides an insight into how the performance of KD varies with the changing temperature of the teacher model. The graph illustrates that applying temperature scaling to teacher improves the performance of the student. Additionally, our experiment shows superior performance even when the teacher is somewhat underconfident as a result of higher temperature settings. This can be attributed to the fact that the true labels act as highly overconfident labels during KD. Consequently, using an underconfident teacher model with these true labels can lead to more balanced probabilities. 


\subsection{Comparison with the state-of-the-art methods}

In our study, we conducted a comprehensive evaluation of our proposed method against previous KD methods~\cite{romero2014fitnets, tian2019crd, mirzadeh2020takd, chen2021reviewkd, zhao2022dkd, jin2023mlld}. The key innovation in our approach is the application of the calibration method to the teacher model before executing KD, which significantly enhances the performance of the student models.
We conducted our experiments across multiple tasks and datasets, which included image classification on CIFAR-100~\cite{krizhevsky2009cifar} and ImageNet~\cite{russakovsky2015imagenet}, as well as object detection on MS-COCO~\cite{lin2014mscoco}. The results of our experiments are presented in Tables \ref{tab:cifar100_same_architecture}, \ref{tab:cifar100_different_architecture}, \ref{tab:imagenet_both}, and \ref{tab:detection} indicate notable performance enhancements when our proposed method is applied.

\subsubsection{Experiment on CIFAR-100 classification} Table~\ref{tab:cifar100_same_architecture} reports the experiment results on the homogenous teacher-student architectures, and Table~\ref{tab:cifar100_different_architecture} reports the experiment results on the heterogeneous teacher-student architectures. With various choices of network pairs, applying our method consistently improves the students' accuracy. Considering the standard deviation of the results and the performance improvement($\Delta$) brought by our method, we observe that applying ours consistently yields comparable or better performance across all models.
It is notable that although it is quite simple, applying the teacher temperature scaling to the previous SOTA method (MLLD + Ours) leads to improved results.
This is quite impressive, considering that the previous SOTA methods in feature distillation (ReviewKD) and logit distillation (MLLD) introduce various kinds of auxiliary networks or additional loss terms. With these experiments, we verify that simply applying temperature scaling draws the better role of the teacher network. 
This demonstrates that previous KD methods were designed without considering calibration errors, and incorporating this aspect into the design could aid in developing future KD methods.
For some pairs, training the student with MLLD + Ours is the only way to surpass its corresponding teacher, whereas the other previous methods fail to do so. It is valuable in the view of model compression in that the ultimate goal of knowledge distillation is to make the student able to replace the teacher.


\subsubsection{Experiment on ImageNet classification} 
In addition to the CIFAR-100 dataset, we extended our experiments to the ImageNet dataset, focusing on ResNet34-ResNet18 and ResNet50-MobileNetV1 architectures, as detailed in Table \ref{tab:imagenet_both}. Our approach demonstrates a marked improvement in performance relative to standard KD, consistent with our findings on the CIFAR-100 dataset. Furthermore, our method exhibits versatility by enhancing the performance of other techniques. For example, when integrated with the MLLD approach, our method lead to better performance.

Additionally, when the MobileNet-V1 was used for the student in Table \ref{tab:imagenet_both}, the abnormal performance drop reveals the harmfulness of using the overconfident logits in the vanilla KD loss. Our method mitigates this by simply applying temperature scaling to the teacher logit.

\begin{table}[t]
\centering
\caption{\textbf{Properties of teacher and student models with calibration method.} The table presents the impact of temperature scaling on various metrics, including $\text{ECE}_{o}$, $\text{ECE}_{u}$, ACE, and ACC, for teachers and students. Bold values denote superior performance in each metric.}
\label{tab:student_property}
\begin{subtable}{0.47\textwidth}
    \centering
    \caption{Teacher: ResNet32x4 and Student: ShuffleNetV1}
    \resizebox{\linewidth}{!}{%
    \begin{tabular}{cccccccc}
    \toprule
    \multicolumn{4}{c}{\textbf{Teacher: ResNet32x4}} & \multicolumn{4}{c}{\textbf{Student: ShuffleNetV1}} \\
    \cmidrule(r){1-4} \cmidrule(l){5-8}
     & $\text{ECE}_{o}$ & $\text{ECE}_{u}$ & ACE & $\text{ECE}_{o}$ & $\text{ECE}_{u}$ & ACE & \text{ACC} \\
    \midrule
    KD   & 9.266 & \textbf{0.035} & \textbf{0.081} & 16.876 & \textbf{0.000} & 0.130 & 72.69 \\
    KD+Ours & \textbf{1.156} & 1.793 & 0.131 & \textbf{8.485} & 0.257 & \textbf{0.101} & \textbf{73.66} \\
    \bottomrule
    \end{tabular}}
\end{subtable}%
\hspace{0.02\textwidth}
\begin{subtable}{0.47\textwidth}
    \centering
    \caption{Teacher: WRN-40-2 and Student: WRN-16-2}
    \resizebox{\linewidth}{!}{%
    \begin{tabular}{cccccccc}
    \toprule
    \multicolumn{4}{c}{\textbf{Teacher: WRN-40-2}} & \multicolumn{4}{c}{\textbf{Student: WRN-16-2}} \\
    \cmidrule(r){1-4} \cmidrule(l){5-8}
     & \(\text{ECE}_{o}\) & \(\text{ECE}_{u}\) & ACE & \(\text{ECE}_{o}\) & \(\text{ECE}_{u}\) & ACE & \text{ACC} \\
    \midrule
    KD   & 11.200 & \textbf{0.019} & 0.100 & 11.125 & \textbf{0.004} & 0.095 & 75.33 \\
    KD+Ours & \textbf{2.444} & 0.574 & \textbf{0.089} & \textbf{2.860} & 0.706 & \textbf{0.089} & \textbf{75.85} \\
    \bottomrule
    \end{tabular}}
\end{subtable}
\end{table}

\begin{table*}[ht]
\centering

\caption{KD experiment results with other calibration methods.}
\label{tab:kd_results}
\begin{tabular}{@{}lccccccccc@{}}
\toprule
Teacher: & VGG19 & VGG13 & VGG16 & ResNet110 & WRN-28-2 & WRN-16-3 & WRN-40-2 & ResNet18 & WRN-16-4 \\ 
\midrule
Teacher Accuracy & 74.10\% & 74.48\% & 74.66\% & 75.22\% & 75.53\% & 75.98\% & 76.55\% & 77.43\% & 77.69\% \\
Mixup Teacher Accuracy & 75.83\% & 76.64\% & 76.49\% & 76.38\% & 76.86\% & 76.42\% & 78.22\% & 79.94\% & 78.19\% \\
\midrule
Vanilla KD & 72.90\% & 73.99\% & 73.39\% & 74.98\% & 75.40\% & 75.37\% & 74.90\% & 74.36\% & 75.33\% \\
\textbf{Temperature Scaling} & 73.37\% & \textbf{74.33\%} & \textbf{73.9\%} & 75.39\% & \textbf{76.12\%} & \textbf{75.95\%} & 75.20\% & \textbf{74.68\%} & 75.95\% \\
\textbf{Vector Scaling} & \textbf{73.47\%} & 74.07\% & 73.63\% & \textbf{75.63\%} & 75.73\% & \textbf{75.95\%} & \textbf{75.26\%} & 74.48\% & \textbf{76.15\%} \\
\textbf{Mixup} & \textbf{74.60\%} & \textbf{74.40\%} & \textbf{73.86\%} & \textbf{76.08\%} & \textbf{76.29\%} & \textbf{76.65\%} & \textbf{75.89\%} & \textbf{74.84\%} & \textbf{76.63\%} \\
\bottomrule
\end{tabular}
\end{table*}

\subsubsection{Experiment on object detection}

We extended our experiments to the object detection task, using the COCO2017 dataset with Faster R-CNN~\cite{ren2015fasterrcnn} teacher-student pairs. As seen in Table~\ref{tab:detection}, applying the calibration method to the teacher model also improved performance in object detection. 
In object detection, feature-based methods demonstrate superior performance as they also convey the spatial information of images. Due to the simplicity of our method, our method can be easily combined with complexly designed feature-based methods. As a result, training students by combining ReviewKD with ours shows consistent performance improvements.

Interestingly, we observed decreased performance for large objects (APl) when adding calibration to the ReviewKD of ResNet101-ResNet18 and ResNet50-MobileNetV2 pairs. We speculate this may be due to our method's influence in reducing the confidence for large objects, while typically large-size objects require higher confidence levels. In contrast, we noted performance improvements when considering medium-size objects (APm) and small-size objects (APs), which are generally more challenging. This leads to an overall enhancement in average AP.

These comprehensive experimental results not only validate the effectiveness of our proposed method but also highlight its potential applicability across various tasks involving probability outputs. By demonstrating the value of taking the calibration error into account in KD, our work paves the way for future research and development in this field, potentially leading to more advanced and efficient KD methodologies.

\subsection{Property of student trained via calibrated teacher}

Table~\ref{tab:student_property} demonstrates the impact of using a teacher with a reduced calibration error on the student's calibration. We observe a significant reduction in the teacher's overconfident calibration error when applying temperature scaling to the teacher. Consequently, the student's accuracy improves, along with reductions in all the metrics of ECE, overconfident ECE, and ACE. Notably, in the case of the ResNet32x4-ShuffleNetV1 pair, although the teacher's ACE increases due to an increase in underconfident error, the student's ACE is actually reduced. This suggests that a slightly underconfident teacher, when combined with highly overconfident true labels, enables the student to learn the actual probabilities more effectively. These experimental results validate that our approach not only enhances the accuracy of the student but also produces a student that is better calibrated. This implies that our approach can yield more reliable models, particularly beneficial when applied in real-world applications.

\subsection{Experiment with other calibration methods}

In this section, we expanded our experiment to include various calibration methods. This additional experiment aimed to address that our contribution is not limited to specific calibration methods but can also be applied to various calibration methods.

We conducted the experiment with various calibration methods and applied these methods to the teacher models. In \mbox{Table~\ref{tab:kd_results}}, we observed that KD performance improves with the application of diverse calibration methods.

We apply vector scaling and mixup. Vector scaling is a widely-used calibration method used in other calibration research like \mbox{\cite{guo2017calibration}} and \mbox{\cite{nixon2019measuring}}. Vector scaling has class size learnable parameters that normalize each output logit. So it has more complexity compared to temperature scaling. Mixup~\mbox{\cite{zhang2017mixup}} is also a widely used augmentation technique that increases the model accuracy and effectively reduces calibration error \mbox{\cite{thulasidasan2019improved, zhang2022mixup_calibration}}. Mixup is applied when training the teacher, and standard KD is performed by using a teacher trained with mixup. The results of our additional experiment consistently align with our finding that applying the calibration method to the teacher can enhance KD performance. This further substantiates our claim. Additionally, it shows that this improvement is not limited to a specific method but also can be applied across various calibration approaches.
In other experiments, we primarily used temperature scaling because vector scaling requires a separate validation set, which could potentially lead to unfair comparisons. Additionally, while mixup reduces calibration error, it also increases teacher accuracy, making it difficult to attribute improvements in knowledge distillation performance solely to reduced calibration error.
Therefore, we chose temperature scaling for its fairness and its ability to reduce calibration error without affecting accuracy. Nevertheless, the experiment results of vector scaling and mixup support our claims and suggest the potential applicability of various calibration methods in KD.
\section{Conclusion}
\label{sec:conclusion}

In this paper, we argue that calibration error is crucial in knowledge distillation. Previously, calibration error was merely considered a supplementary metric indicating a model's reliability or robustness. However, by demonstrating that calibration error also impacts the performance of the student in KD, we have significantly expanded the role of calibration error in the realm of deep learning.

Through this paper, we empirically show a strong correlation between calibration error and KD performance. This leads us to propose that the calibration errors of teachers should be considered a new design criterion. We have shown the validity of this criterion through various experiments.

Furthermore, we discovered that by applying a simple calibration method to reduce calibration error, substantial performance improvements could be achieved in the standard KD method. This approach is also applicable to existing state-of-the-art methods, demonstrating the possibility of additional performance enhancements through its application.

Overall, our findings highlight the importance of teacher calibration error in knowledge distillation and provide a foundation for further advancements in KD methods.

\textbf{Limitation.}
While this paper empirically demonstrates the significant role of overconfident calibration error in KD, it is important to note that our study is limited in scope to logit distillation methods. Feature distillation is also an important technique widely employed in deep learning applications. We have not explored the impact of calibration error on feature distillation methods, thereby indicating a need for further research in this area.


\bibliographystyle{IEEEtran}
\bibliography{ref}

\EOD

\end{document}